# Approximations for Decision Making in the Dempster-Shafer Theory of Evidence

**Mathias Bauer**
German Research Center for Artificial Intelligence (DFKI)
Stuhlsatzenhausweg 3
66123 Saarbrücken, GERMANY
email: bauer@dfki.uni-sb.de

## Abstract

The computational complexity of reasoning within the Dempster-Shafer theory of evidence is one of the main points of criticism this formalism has to face. To overcome this difficulty various approximation algorithms have been suggested that aim at reducing the number of focal elements in the belief functions involved. Besides introducing a new algorithm using this method, this paper describes an empirical study that examines the appropriateness of these approximation procedures in decision making situations. It presents the empirical findings and discusses the various tradeoffs that have to be taken into account when actually applying one of these methods.

## 1 Introduction

The complexity of the computations that have to be carried out in the Dempster-Shafer theory of evidence (DST) [Dempster, 1967; Shafer, 1976] is one of the main points of criticism this formalism has to face. In fact, [Orponen, 1990] shows that the combination of two basic probability assignments (bpa's) using Dempster's rule is #P-complete.[1] To overcome this difficulty various approximation methods have been suggested. While some of them concentrate on the processing of particular types of information only—e.g. the methods proposed in [Barnett, 1981] and [Gordon & Shortliffe, 1985]—others try to attack this problem for arbitrary bodies of evidence.

The number of *focal elements* in the belief functions under consideration heavily influences the computational complexity of combining various independent pieces of evidence. As a consequence, [Voorbraak, 1989], [Tessem, 1993], and [Lowrance, Garvey, & Strat, 1986] suggest approximations that aim at reducing this influence factor by removing focal elements and/or redistributing the corresponding numerical values.

An empirical study presented in [Tessem, 1993] investigates the quality of the approximations so obtained by considering the average maximal deviation from function values that are relevant to decision making. A second criterion to assess the quality of an approximation is its ability to induce some "structure" on the given information, thus allowing an adequate presentation of the essentials of its contents to a user.

While this study focused on the *quantitative* aspects of decision making only, the *quality* of a decision is more important in many applications. That is, whenever the numbers only serve as a means for finding the best candidate among a number of alternatives, the numerical deviation from the actual values is of secondary interest only. As a consequence the empirical study presented in this paper will consider both numerical and qualitative aspects when evaluating the appropriateness of various approximation algorithms for efficient decision making. Building upon the methodology described in [Tessem, 1993] new measures capturing the deviation from an optimal choice are introduced and the actual output size of the methods under consideration is recorded, thus providing the basis for discussing the various tradeoffs associated with the application of one of the algorithms presented.

In addition to investigating the properties of various algorithms known from the literature, this paper also introduces a new approximation method—the D1 algorithm—and compares its performance with that of the other candidates.

The rest of this paper is organized as follows. Section 2 reviews the basic notions of DST and motivates the need for approximate computations. Section 3 discusses four approximation procedures including the new D1 algorithm before Section 4 describes the testbed used and interprets the empirical results. Finally, Section 5 concludes with a brief summary and a reference to related work. An appendix contains the graphs illustrating the empirical findings.

## 2 The Need for Approximations

The basic carrier of information in the Dempster-Shafer theory of evidence (DST) is a *basic probability assignment*

---

[1]The class #P is a functional analogue of the class NP of decision problems.

(bpa),[2] a function $m : 2^\Theta \to [0, 1]$ that assigns a numerical value to each subset of a given frame of discernment $\Theta$ and satisfies the properties

- $m(\emptyset) = 0$ and

- $\sum_{A \subseteq \Theta} m(A) = 1$.

Subsets $A \subseteq \Theta$ with $m(A) > 0$ are called the *focal elements* of $m$. Two bpa's $m_1$ and $m_2$ can be combined using *Dempster's rule* [Dempster, 1967]:

$$(m_1 \oplus m_2)(A) = \frac{\displaystyle\sum_{A_1 \cap A_2 = A} m_1(A_1) \, m_2(A_2)}{\displaystyle\sum_{A_1 \cap A_2 \neq \emptyset} m_1(A_1) \, m_2(A_2)}$$

Dealing with bpa's brings about both representational and computational complexity problems that are discussed in [Orponen, 1990] where the combination of various pieces of evidence using Dempster's rule is shown to be #P-complete. Given a frame of discernment of size $|\Theta| = N$, a bpa $m$ can have up to $2^N$ focal elements all of which have to be represented explicitly to capture the complete information encoded in $m$. Furthermore, the combination of two bpa's requires computation of up to $2^{N+1}$ intersections.

Reducing the number of focal elements of the bpa's under consideration *while retaining the essence of the information they represent* is thus a viable way to overcome these problems. The next section will present a number of approximation algorithms all of which make use of this basic idea.

## 3  Approximation Algorithms

This section reviews various approximation algorithms known from the literature, introduces the new D1 algorithm, and discusses their basic characteristics including their respective computational complexity and the size of their output in terms of the number of focal elements left after approximation. Combined with the empirical results described in Section 4.2, this will form the basis for the evaluation of the various tradeoffs associated with the application of a particular approximation method in decision making situations.

### 3.1  The Bayesian Approximation

The *Bayesian approximation* [Voorbraak, 1989] reduces a given bpa $m$ to a (discrete) probability distribution, i.e. only singleton subsets of the frame of discernment are allowed to be focal elements of the approximated version $\underline{m}$ of $m$:

$$\underline{m}(A) = \begin{cases} \dfrac{\displaystyle\sum_{B:A\subseteq B} m(B)}{\displaystyle\sum_{C:C\subseteq\Theta} m(C) \cdot |C|}, & |A| = 1 \\[6pt] 0, & \text{otherwise} \end{cases}$$

As a consequence $\underline{m}$ can have at most $|\Theta|$ focal elements. Given a bpa $m$ with $n$ focal elements, its Bayesian approximation can be computed in time $O(n \cdot |\Theta|)$.

**Remarks:**

1. $m_1 \oplus m_2 = \underline{m_1} \oplus \underline{m_2}$, i.e. the order of combination and approximation does not influence the final result. Among all the algorithms taken into consideration in this paper, the Bayesian approximation is the only one with this property.

2. If $m_1$ is a Bayesian bpa (i.e. $m_1$ has only singleton focal elements), then $\underline{m_1 \oplus m_2} = m_1 \oplus m_2$.

The following example will be used throughout the rest of this section to illustrate the effects of the various approximation methods described.

**Example:**
Let $m$ be a bpa over the frame of discernment $\Theta = \{a, b, c, d, e\}$ with the values

$$m(A) = \begin{cases} 0.50, & A = \{a, b\} \\ 0.30, & A = \{a, c, d\} \\ 0.10, & A = \{c\} \\ 0.05, & A = \{c, d\} \\ 0.05, & A = \{d, e\} \\ 0, & \text{otherwise} \end{cases} \tag{1}$$

Applying the Bayesian approximation to $m$ yields the following result

$$\underline{m}(A) \approx \begin{cases} 0.360, & A = \{a\} \\ 0.230, & A = \{b\} \\ 0.205, & A = \{c\} \\ 0.180, & A = \{d\} \\ 0.023, & A = \{e\} \\ 0, & \text{otherwise} \end{cases}$$

This example demonstrates that the application of this approximation method is not reasonable in cases where the number of focal elements of the input bpa $m$ itself is $\leq |\Theta|$.

### 3.2  The $k$-$l$-$x$ Method

The basic idea of this approximation introduced in [Tessem, 1993] is to incorporate only the highest valued focal elements of the original bpa into the approximation $m_{klx}$ as long as at least $k$ and at most $l$ focal elements remain and the sum of the $m$-values of the remaining focal elements is at least $1 - x$ where $x \in [0, 1[$. Finally the remaining values are normalized in order to guarantee the basic properties of a bpa (see Section 2). The exact definition of this algorithm is given in Figure 1. For a bpa $m$ with $n$ focal elements the approximation $m_{klx}$ can be computed in time $O(n \cdot log\, n)$.

**Example:**
For the bpa $m$ as given in (1) and the values $k = 2$, $l = 3$,

**begin** $k\text{-}l\text{-}x$–approximation$(m, k, l, x)$
  sort the subsets of $\Theta$ w.r.t. their $m$-values
  $totalmass := 0$   % total mass of $m_{klx}$ so far
  $f := 0$       % number of focal elements of $m_{klx}$
  **while** $m$ contains focal elements **and**
      $(f \leq l)$ **and**
      $((f < k)$ **or** $(totalmass < 1 - x))$
      **do** add next focal element $A$ of $m$ to $m_{klx}$
        $f := f + 1$
        $totalmass := totalmass + m(A)$
      **od**
  normalize the values of $m_{klx}$
**end**

Figure 1: The $k\text{-}l\text{-}x$ Approximation.

and $x = 0.1$, the following results are obtained.

$$m_{klx}(A) = \begin{cases} 0.\overline{5}, & A = \{a, b\} \\ 0.\overline{3}, & A = \{a, c, d\} \\ 0.\overline{1}, & A = \{c\} \\ 0, & \text{otherwise} \end{cases}$$

Removing $\{c, d\}$ and $\{d, e\}$—each with a value of 0.05—from $m$ was sufficient to satisfy both the constraints concerning the number of focal elements left and the numerical mass deleted. In this case, normalizing the remaining values is equivalent to dividing them by 0.9—the value of $1 - (m(\{c, d\}) + m(\{d, e\}))$.

### 3.3 Summarization

Similar to the $k\text{-}l\text{-}x$ procedure, the summarization method [Lowrance, Garvey, & Strat, 1986] leaves the best-valued focal elements of the bpa under consideration unchanged. The numerical values of the remaining focal elements are accumulated and assigned to the set-theoretic union of the corresponding subsets of $\Theta$.

Let $k$ be the number of focal elements to be contained in the approximation $m_S$ of a given bpa $m$. $M$ denotes the set of the $k - 1$ subsets of $\Theta$ with the highest values in $m$. Then $m_S$ is given by

$$m_S(A) = \begin{cases} m(A), & A \in M \\ \sum_{\substack{A' \subseteq A \\ A' \notin M}} m(A'), & A = A_0 \\ 0, & \text{otherwise} \end{cases}$$

where $A_0$ is uniquely determined by

$$A_0 = \bigcup_{\substack{A' \notin M \\ m(A') > 0}} A'$$

The summarization of a bpa with $n$ focal elements can be computed in time $O(n)$.

**Example:**

For the bpa $m$ from (1) and $k = 3$, $m_S$ has the following values.

$$m_S(A) = \begin{cases} 0.5, & A = \{a, b\} \\ 0.3, & A = \{a, c, d\} \\ 0.2, & A = \{c, d, e\} \\ 0, & \text{otherwise} \end{cases}$$

In this case $A_0$ corresponds to $\{c, d, e\}$, the set $M$ of the $k - 1$ best-valued focal elements of $m$ contains $\{a, b\}$ and $\{a, c, d\}$.

### 3.4 The D1 Approximation

Let $m$ be a bpa to be approximated. Again $k$ is the desired number of focal elements of the approximated bpa $m_D$. Furthermore, let $M^+$ denote the set of the $k - 1$ focal elements of $m$ with the highest values, $M^-$ the set containing all other focal elements of $m$:

$$M^+ = \{ A_1, ..., A_{k-1} \subseteq \Theta \mid$$
$$\forall A \notin M^+ : m(A_i) \geq m(A), i = 1, ..., k - 1 \}$$
$$M^- = \{A \subseteq \Theta \mid m(A) > 0, A \notin M^+\}$$

The crucial idea of the D1 algorithm is to keep all the members of $M^+$ as focal elements of $m_D$ and to distribute the numerical values of the elements in $M^-$ among them. This distribution works as follows. Given a focal element $A \in M^-$ of $m$, the collection $M_A$ of supersets of $A$ in $M^+$ is computed. The value $m(A)$ is dispensed uniformly among the (set-theoretically) smallest members of $M_A$.

In case $M_A$ is empty, i.e. $M^+$ contains no superset of $A$, the set $M'_A$ is constructed:

$$M'_A = \{B \in M^+ \mid |B| \geq |A|, B \cap A \neq \emptyset\}.$$

Again $m(A)$ is shared among the smallest members of $M'_A$. The exact value to be assigned to a focal element depends on the size of its intersection with $A$. This procedure is invoked recursively until all of $m(A)$ could be assigned to the members of $M^+$ or the set $M'_A$ becomes empty. In this case the remaining value is assigned to $\Theta$ which thus becomes a focal element of $m_D$. The D1 algorithm is described in detail in Figure 2.

The approximation of a bpa with $n$ focal elements can be computed in time $O(k \cdot (n - k))$.

**Remarks:**

1. This approximation method is *conservative* in that the numerical values of focal elements to be removed are exclusively assigned to (set-theoretically) *larger* subsets of $\Theta$, i.e. the approximated version $m_D$ represents information that is *less specific* than the one originally encoded in $m$.

2. The distribution of the value $m(A)$ of a focal element $A \in M^-$ among its supersets is intended to bring about minimal deviations in the pignistic probability induced by $m_D$ (see Section 4). This effect is even increased **begin** D1_approximation$(m, M^+, M^-)$
    **forall** $A \in M^+$
    **do** $m_D(A) := m(A)$ **od**
    $m_D(\Theta) := m(\Theta)$
    **forall** $A \in M^-$
    **do** $distribute(A, m(A), |A|)$ **od**
**end**

**procedure** $distribute(A : set, val : real, limit : integer)$
$M_A := \{B \in M^+ \mid A \subset B\}$
**if** $M_A \neq \emptyset$
**then** % *case 1:* $M^+$ *contains supersets of* $A$

    $\hat{M}_A := \{B \in M_A \mid |B| \text{ minimal in } M_A\}$
    **forall** $B \in \hat{M}_A$
    **do**  % *uniform distribution of* $m(A)$ *among*
        *the members of* $\hat{M}_A$:

      $m_D(B) := m_D(B) + (1/|\hat{M}_A|) \cdot val$

    **od**

**else** % *case 2:* $M^+$ *contains no supersets of* $A$

    $M'_A := \{B \in M^+ \mid |B| \geq limit, A \cap B \neq \emptyset\}$
    **if** $M'_A = \emptyset$
    **then** $m_D(\Theta) := m_D(\Theta) + val$
    **else** $hlp := 0$  % *part of* $m(A)$ *already distributed*

      $\hat{M}_A := \{B \in M'_A \mid |B| \text{ minimal in } M'_A\}$

      % *let* $\hat{M}_A = \{B_1, ..., B_l\}$

      % *determine portion of elements of* $A$
        *occurring in the various* $B_i$:

      $ratio := \dfrac{|\cup_{i=1}^l B_i \cap A|}{|A|}$

      % *compute total number of "shares"*:

      $number := \sum_{i=1}^l |B_i \cap A|$

      **forall** $B \in \hat{M}_A$
      **do** $share := (|B \cap A|/number) \cdot ratio \cdot val$
        $m_D(B) := m_D(B) + share$
        $hlp := hlp + share$
      **od**
      **if** $ratio < 1$
      **then** $A_{rest} := A \setminus \cup_{i=1}^l B_i$  % *remainder of* $A$
        **if** $A_{rest} \neq \emptyset$
        **then** $distribute(A_{rest}, val - hlp, limit)$
        **fi**
      **fi**
    **fi**
**fi**

Figure 2: The Algorithm D1.

---

by the concentration on the smallest possible supersets of $A$.

**Example:**
The D1 algorithm yields the following results for the bpa $m$ from (1) and $k = 3$.

$$m_D(A) = \begin{cases} 0.5, & A = \{a, b\} \\ 0.475, & A = \{a, c, d\} \\ 0.025, & A = \{a, b, c, d, e\} \\ 0, & \text{otherwise} \end{cases}$$

Obviously, $\{c\}$, $\{c, d\}$, and $\{d, e\}$ belong to $M^-$. The focal element $\{a, b\}$ has an empty intersection with all of these sets. As a consequence its value remains unchanged. $\{a, c, d\}$ is the only superset of $\{c\}$ and $\{c, d\}$ in $M^+$, such that its value is increased by $0.10 + 0.05$. Furthermore it covers half of the elements of $\{d, e\}$ which adds up another $0.05/2$ to the $m_D$ value of $\{a, c, d\}$. The rest is assigned to $\Theta = \{a, b, c, d, e\}$.

## 4   Empirical Tests and Results

A number of empirical tests were performed with the aim to investigate the appropriateness of the various approximation algorithms presented in the last section in decision making situations. Section 4.1 will describe the test environment and the error measures used to quantify the induced deviations from an optimal decision, before Section 4.2 gives an overview of the results and discusses the tradeoffs associated with the application of a particular algorithm.

### 4.1   The Testbed

To arrive at comparable results, many parameters of the testbed were taken from [Tessem, 1993]. A frame of discernment of size $|\Theta| = 32$ was assumed as the basis for all the tests. For each algorithm random bpa's with 8 focal elements were generated and 5 combinations using Dempster's rule were computed.[3] This was repeated more than 1000 times for each candidate. After each combination the results were approximated and evaluated w.r.t. the error measures described below.

Bpa's are generated using exponentially distributed random numbers $X$ and $Y$ (with identical distributions) and the following algorithm:

    $rest := 1$
    **for** $i := 1$ **to** 7
    **do** generate random number $X$
        randomly generate $A \subseteq \Theta$
        $m(A) := P(Y \leq X) \cdot rest$
        $rest := rest - m(A)$
    **od**
    generate "new" $A \subseteq \Theta$
    $m(A) := rest$

---

[3][Tessem, 1993] also investigates bpa's with 2 and 4 focal elements.

Table 1: Quantitative Comparison of the Approximation Algorithms.

|  | average | | minimum | | maximum | |
|---|---|---|---|---|---|---|
|  | orig | approx | orig | approx | orig | approx |
| D1_8 | 1624 | 7.90 | 213 | 7 | 7058 | 8 |
| D1_30 | 1616 | 29.93 | 208 | 29 | 4590 | 30 |
| Summ_8 | 1680 | 8.00 | 209 | 8 | 5859 | 8 |
| Summ_30 | 1589 | 30.00 | 188 | 30 | 4404 | 30 |
| Bayes | 1632 | 29.32 | 217 | 23 | 5806 | 31 |
| klx_01 | 1616 | 440.02 | 234 | 3 | 4447 | 1849 |
| klx_30 | 1448 | 28.63 | 179 | 13 | 2763 | 30 |

This way a uniform distribution of numerical values among the focal elements is avoided, i.e. the random bpa's are closer to "realistic" data where the information given supports the various alternatives to different degrees.

For the experiments the following parameter values were chosen. For D1 and the summarization algorithm the instantiations with $k = 8$ and $k = 30$—denoted by D1_8, D1_30, Summ_8, and Summ_30, resp.—were considered. For the $k$-$l$-$x$ method the parameter sets were $k = 1, l = \infty, x = 0.01^4$ and $k = 1, l = 30, x = 1.0$. The corresponding instantiations of this algorithm are denoted by klx_01 and klx_30, resp. Recall that the Bayesian approximation algorithm is not parameterized.

All the error measures used are based on the *pignistic probability* $P_0$ induced by a bpa that can be considered the standard function for decision making in DST [Smets, 1988]. It is characterized by

$$P_0(\{x\}) = \sum_{A : x \in A \subseteq \Theta} \frac{m(A)}{|A|}.$$

To keep the results comparable to those obtained in [Tessem, 1993], the first error measure is identical to the one used there. It quantifies the maximal deviation in the pignistic probability induced by an approximated bpa. Let $P_0$ be the pignistic probability induced by the original version of a bpa $m$, $P_{app}$ the one induced by its approximation $m'$ (no matter which algorithm was applied). Then the error measure is defined as

$$\text{Error1}(m') = \max_{A \subseteq \Theta} |P_0(A) - P_{app}(A)|$$

This measure, however, does not reflect the *quality* of a decision based on $P_{app}$. To do so, additional error measures are introduced. Let $x_0, x_{app} \in \Theta$ be characterized by the following properties:

$$P_0(\{x_0\}) = \max_{x \in \Theta} P_0(\{x\}).$$

$$P_{app}(\{x_{app}\}) = \max_{x \in \Theta} P_{app}(\{x\}),$$

i.e. $x_0$ ($x_{app}$) is the best choice among all alternatives from $\Theta$ given $P_0$ ($P_{app}$). In this situation, the quantities

$$\text{Error2}(m') = |\{ x \mid P_{app}(\{x\}) > P_{app}(\{x_0\}) \}|$$

$$\text{Error3}(m') = |\{ x \mid P_0(\{x\}) > P_0(\{x_{app}\}) \}|$$

count the numbers of alternatives with a higher $P_{app}$ value than $x_0$ and a higher $P_0$ value than $x_{app}$, resp. This means they reflect the rankings of $x_0$ (the choice that should be made) on the basis of $P_{app}$ and of $x_{app}$ (the choice actually made) on the basis of $P_0$.

For example, given the concrete values of $\text{Error2}(m') = 1$ and $\text{Error3}(m') = 2$ for some approximated bpa $m'$ means that the third best choice is made while the actual optimum is considered the second best alternative only.

Error3 is particularly important for the assessment of an approximation method w.r.t. decision making as it directly represents the quality of decision $x_{app}$. A value of 0 indicates that the approximation yields the same optimal choice as the original information. Note that in this case Error2 also assumes the value 0.

### 4.2 The Results

For a fair comparison of the various approximation algorithms including a characterization of the respective trade-offs associated with their application, it is not sufficient to merely consider these error measures. In addition the size of their output has to be taken into account to estimate the gain in runtime. Table 1 summarizes the average, minimum, and maximum number of focal elements both in the original data and their approximations for each candidate algorithm after the fifth combination.

**Remarks:**
1. For technical reasons the algorithms were not run with identical test data. However, the statistical data in

---

[4]The same values as in [Tessem, 1993].

Table 1 indicate that the average problem size was approximately equal for all candidates.

2. The extreme number of 1849 focal elements was produced by klx_01 when the original bpa contained 3843 focal elements. The approximations generated by this algorithm exceeded 1000 focal elements in more than 7 % of all cases.

The Figures 3, 4, and 5 at the end of this paper (see Appendix A) depict the averaged results for the error measures described above.

In all three cases klx_01 reaches the best values. This is due to the fact that only relatively few focal elements with extremely low values are removed from the input data. The average size of a bpa approximated with this method is 440 focal elements, the maximum is even more than 1800 (see Table 1). As a consequence the gain in runtime is the least among all candidates—taking into account both the time to compute the combinations and the approximation itself. Restricting the number of focal elements as in klx_30 improves this aspect but induces significant deviations in Error1.

Compared to algorithms with similar output size, both instantiations of the summarization yield the worst values for all error measures.

The Bayesian approximation is the only one with improving values after several combinations. In [Tessem, 1993] this is explained by the fact that the result of combining several bpa's becomes more and more specific and thus approximates a probability distribution. As a consequence this algorithm yields its worst results whenever decisions have to be made on the basis of little evidence only. After 4 to 5 combinations it reaches the level of D1_30.

The errors induced by D1 are relatively small even for the instantiation with 8 focal elements only (this is particularly true for Error2 and Error3). This facilitates decision making in extremely time critical applications. The D1_30 instantiation yields the best values of all candidates (with the exception of klx_01) during the first combinations. After the 4th to 5th step the Bayesian approximation starts to produce slightly better values.

What are the consequences for the selection of a particular algorithm in a given application? If the runtime behavior is not too critical, klx_01 or similar instantiations of the $k$-$l$-$x$ algorithm are the best choice.[5] If only up to $|\Theta|$ focal elements are acceptable, D1_30 and the Bayesian approximation yield the best results. The former is advantageous during the first combinations, while the latter is preferable during longer cycles. If time restrictions allow only very coarse approximations, D1 outperforms the competitors as the example of D1_8 shows.

**Remark:** The positive results for the Bayesian approximation do not come for free. It collapses the belief intervals $[Bel(.), Pl(.)]$ into point values. As a consequence, the ability of DST to explicitly represent and deal with partial ignorance is lost.

# 5  Conclusion

The results presented in the preceding section clearly show that the "best" approximation algorithm with respect to decision making does not exist. Instead the tradeoffs between the number of focal elements remaining, the complexity of computing the approximation itself, and the quality of the decisions made have to be taken into account. However, it can be stated that the $k$-$l$-$x$ algorithm, D1, and the Bayesian approximation yield definitely better results than the summarization does. Given a particular application, a ranking of these three alternatives can be established on the basis of the discussion in Section 4.2.

This also applies to the *consonant approximation* [Dubois & Prade, 1990]. The test results of this approximation method were significantly worse than those of all other algorithms examined (as was already observed in [Tessem, 1993]). As a consequence this candidate was not further discussed in this paper.

In addition to the approximation methods presented here, several *Monte-Carlo algorithms* were suggested to make reasoning in the DST computationally feasible (e.g. [Wilson, 1991; Kreinovich *et al.*, 1994]). Their application involves making a number of random choices on the basis of some probability distribution. Trying to arrive at a correct result with a certain probability, their usage can be motivated by the common observation that even the initial input data given by an expert are reliable only to a certain degree such that it is justifiable to have an algorithm the output of which is slightly incorrect in a limited number of cases [Kreinovich *et al.*, 1994]. Future work will include the examination of the properties of such algorithms in practical applications.

#### Acknowledgements

This work was supported by the BMBF under grant ITW 9404 0 as part of the RAP project.

I'd like to thank my colleagues Dietmar Dengler and Harald Feibel for their comments on earlier versions of this paper and Patrick Brandmeier for his support in implementing the testbed and running the experiments.

# A  The Error Graphs

This appendix contains the graphs depicting the average values of the measures Error1 through Error3 defined in Section 4.1 for each of the approximation methods discussed in Section 3 and the corresponding instantiations of their parameters mentioned above.

---

[5]Further experiments showed that all 3 error measures grow approximately linearly with the value of $x$.

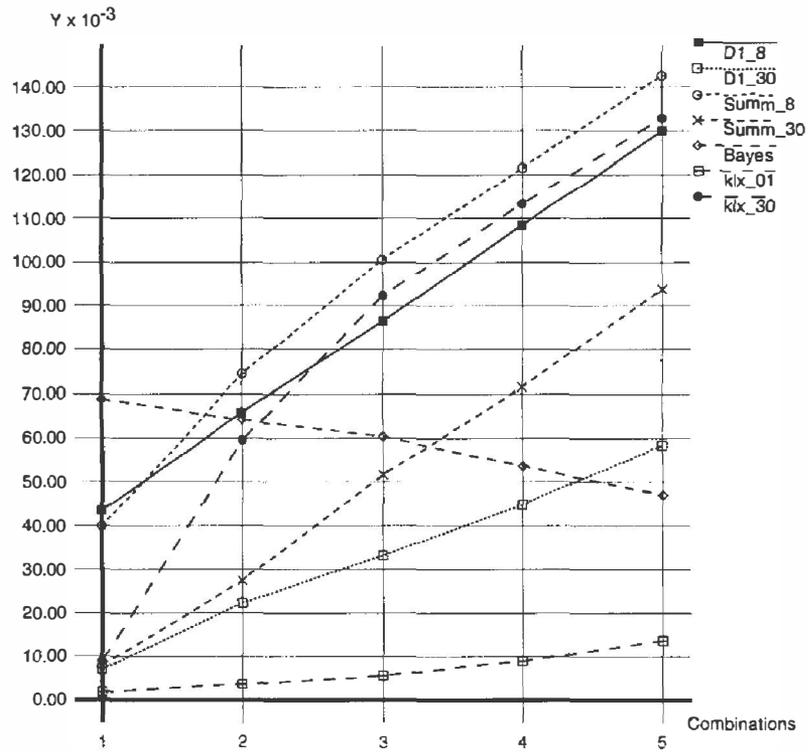

Figure 3: Results for Error1.

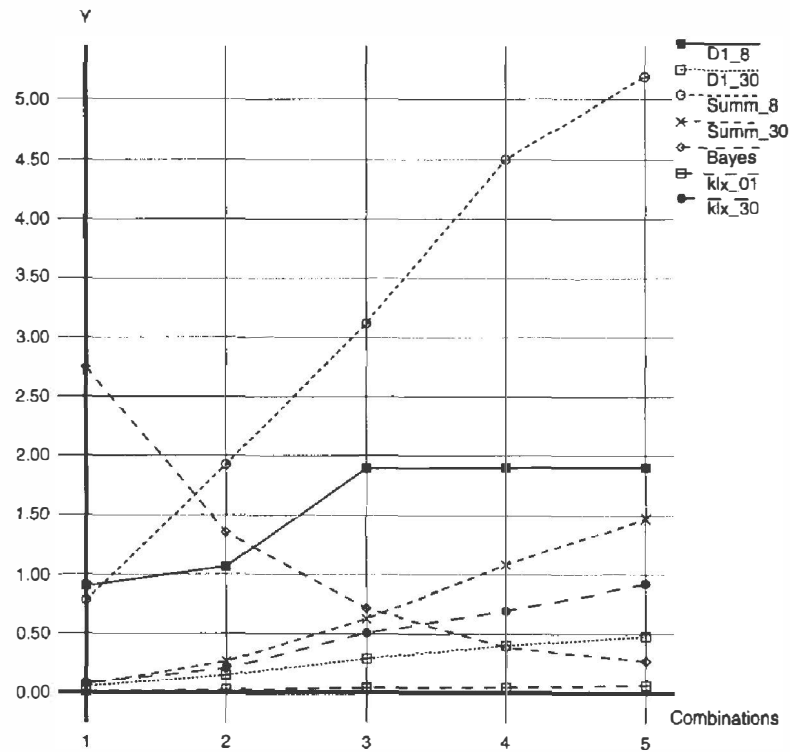

Figure 4: Results for Error2.

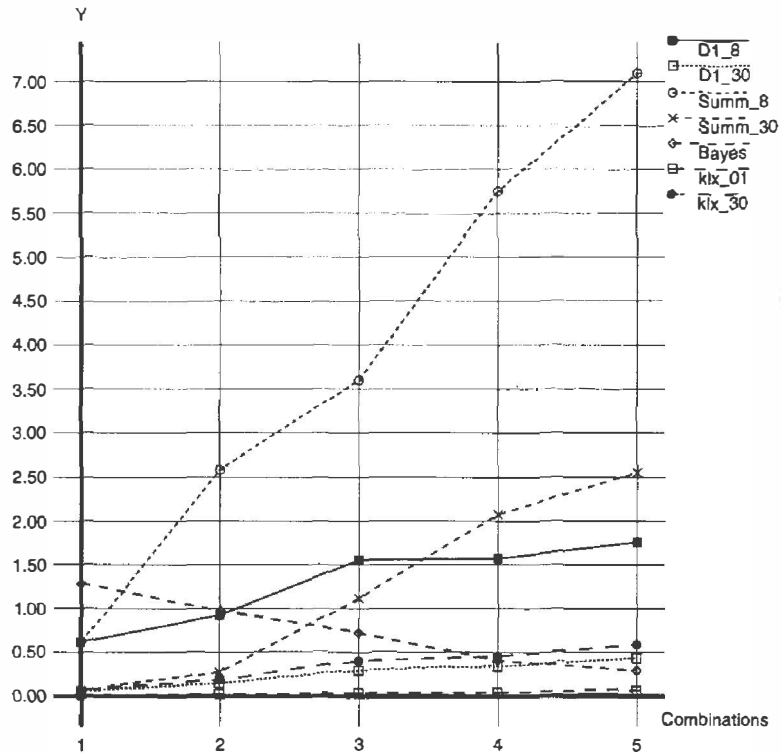

Figure 5: Results for Error3.